\documentclass[runningheads]{llncs}
\usepackage{graphicx}
\usepackage{booktabs}
\usepackage{multirow}
\usepackage{threeparttable}
\usepackage{tabularx}
\usepackage{makecell}
\usepackage{xcolor}
\usepackage{adjustbox}
\usepackage{amssymb}
\usepackage{booktabs}
\usepackage{multirow}
\usepackage{makecell}
\usepackage{adjustbox}
\definecolor{groupgray}{RGB}{245,247,250}
\definecolor{headgray}{RGB}{250,251,253}
\definecolor{commblue}{RGB}{70,110,180}
\definecolor{osgreen}{RGB}{70,130,90}
\definecolor{oursblue}{RGB}{232,239,250}
\usepackage[table]{xcolor}
\usepackage[table]{xcolor}
\usepackage{pifont}
\newcommand{\cmark}{\ding{51}}
\newcommand{\xmark}{\ding{55}}
\newcommand{\na}{--}
\usepackage{graphicx}
\usepackage{amsmath}
\usepackage{booktabs}
\usepackage{amssymb}
\usepackage{array}
\usepackage{tabularx}
\usepackage{xcolor}
\usepackage{colortbl}
\newcommand{\best}[1]{\cellcolor{commblue!18}#1}
\newcommand{\second}[1]{\cellcolor{osgreen!18}#1}
\providecommand{\cmark}{\raisebox{0.12ex}{\small$\checkmark$}}
\providecommand{\xmark}{\raisebox{0.12ex}{\small$\times$}}
\providecommand{\na}{\textcolor{gray}{--}}
\usepackage{float}
\usepackage[hidelinks]{hyperref}

\begin{document}
\title{VGIF-Score: Interpretable and Diagnostic Evaluation of Spatio-Temporal Instruction Following in Video Generation}
\titlerunning{VGIF-Score}
%
\author{Songyu Xu\inst{1} \and Xin Wang\inst{1} \and Qiang Chen\inst{2} \and Xinran Wang\inst{1} \and Muxi Diao\inst{1} \and Yuxuan Zhang\inst{1} \and Kongming Liang\inst{1} \and Rui Lin\inst{2}\thanks{Corresponding author} \and Zhanyu Ma\inst{1}}
\institute{Beijing University of Posts and Telecommunications, Beijing, China \and China Telecommunications Group Co., Ltd., Beijing, China \\\email{linrui@chinatelecom.cn} }

\authorrunning{S. Xu et al.}
%
\maketitle           
\begin{abstract} Recent video generation models (VGMs) have made substantial progress in visual fidelity, yet their ability to follow long, compositional instructions remains insufficiently evaluated. Existing evaluation protocols often rely on prompts that are short and semantically shallow, with limited atomic constraints and weak spatio-temporal dependencies. They also frequently depend on costly human evaluation or handcrafted vision pipelines, while providing little diagnostic insight into which instruction constraints succeed or fail. To address this gap, we propose \textbf{VGIF-Score}, a highly automated and interpretable framework for evaluating instruction following in video generation. VGIF-Score consists of two complementary components: an \emph{objective completion branch} that parses prompts into a Spatio-Temporal Directed Acyclic Graph (ST-DAG) and performs dependency-aware QA with short-circuit diagnostics, and a \emph{subjective satisfaction branch} that uses instruction-conditioned AutoRubric to assess cinematography, visual purity, motion smoothness, and physics adherence. Together, these components produce a unified score that captures both objective completion and perceptual satisfaction. We instantiate this framework on \textbf{VGIF-Bench}, a benchmark of 223 long, structurally entangled prompts paired with approximately 4.3K fine-grained evaluation items. Experiments on 14 proprietary and open-source VGMs across more than 3K generated videos show that VGIF-Score provides reliable, interpretable, and diagnostically useful evaluation of video generation instruction following. The code will be available at \url{https://github.com/PRIS-CV/VGIF-SCORE}.

\end{abstract}

\section{Introduction}

Video Generation Models (VGMs) have rapidly progressed from early generative paradigms such as VAEs, GANs, and autoregressive modeling~\cite{VAE,goodfellow2014generative,yan2021videogpt} to diffusion and diffusion-transformer architectures~\cite{ho2022video,chen2026seeing,gao2026toward,gao2023self,qin2025multimodal,qin2026increfa,wang2026udm}. Recent large-scale systems~\cite{wan2025wan,team2025kling} can produce visually faithful and increasingly cinematic videos through improved spatio-temporal modeling. Despite this progress, instruction following remains a key bottleneck. A model may handle ``a girl walking in a park,'' but struggle with ``a girl in a red coat drops a glass, the glass shatters, and a nearby dog turns toward the sound''---a prompt that tests whether the model captures the logic of an event sequence, not merely individual visual phenomena. This gap reveals the need to evaluate spatio-temporal instruction following beyond visual plausibility.

Existing evaluation protocols, however, remain insufficient. Traditional metrics such as FVD~\cite{unterthiner2018towardsFVD} and CLIP-based scores~\cite{clipscore} capture only low-level similarity or coarse semantics. Recent benchmarks broaden the landscape through multi-dimensional evaluation~\cite{vbench,huang2025vbench++,vbench2.0}, compositional or physical reasoning tests~\cite{sun2025t2vcompbench,chen2026t2vworldbench}, and human-aligned or MLLM-based judging~\cite{evalcrafter}. Yet two fundamental limitations persist. First, prompts are typically short and semantically shallow. Even benchmarks with longer prompts~\cite{huang2025vbench++} often evaluate coarse dimensions rather than dependency-aware execution. We argue that instruction difficulty is governed not by length alone, but by \emph{compositional depth}---the number of atomic constraints and the dependencies among them---where existing benchmarks remain shallow (Table~\ref{tab:benchmark_comparison}). Second, scoring is largely aggregate, offering little diagnostic insight into \emph{which} constraints fail and \emph{how} failures propagate through causal chains.

\begin{figure*}[t]
    \centering
    \includegraphics[width=\textwidth]{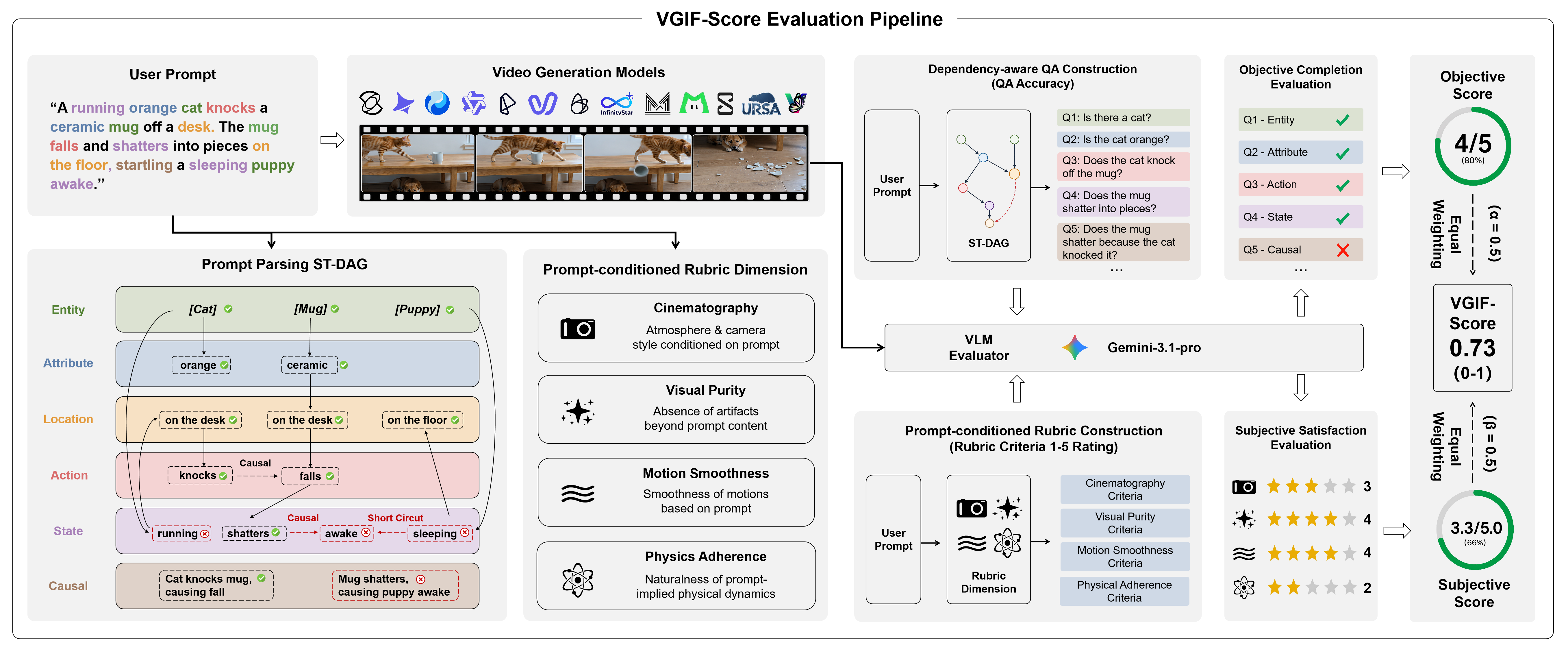}
    \caption{Overview of VGIF-Score. The framework evaluates spatio-temporal instruction following via objective QA-based scoring and subjective rubric-based assessment}
    \label{fig:vgif_pipeline}
\end{figure*}

To address these limitations, we propose \textbf{VGIF-Score}, a fine-grained and automated framework for evaluating instruction following in video generation (Figure~\ref{fig:vgif_pipeline}). Inspired by Davidsonian Scene Graphs~\cite{dsg}, we decompose each prompt into a \textbf{Spatio-Temporal Directed Acyclic Graph (ST-DAG)} of atomic semantic units---entities, attributes, locations, actions, states, and causal relations---connected by explicit dependency edges. From this graph, we derive dependency-aware QA pairs with a short-circuit mechanism that propagates failures along the dependency structure, and complement them with an instruction-conditioned \textbf{AutoRubric}~\cite{autorubric} that assesses cinematography, visual purity, motion smoothness, and physics adherence. We instantiate this framework on \textbf{VGIF-Bench}, a diagnostic benchmark of \textbf{223 long-form, dependency-rich prompts} and approximately \textbf{4.3K fine-grained evaluation items}. Experiments on 14 open-source and proprietary VGMs across more than 3K generated videos show that VGIF-Score provides reliable and interpretable evaluation, and reveals two systematic failure modes: weak \emph{causal instruction following} and strong sensitivity to \emph{depencey-depth} and \emph{prompt position}.

\begin{table*}[t]
\caption{Comparison of representative video-generation evaluation benchmarks.}
\label{tab:benchmark_comparison}
\centering
\scriptsize
\renewcommand{\arraystretch}{0.96}
\setlength{\tabcolsep}{4.2pt}
\begin{tabular*}{\textwidth}{@{\extracolsep{\fill}}lcccccccccc@{}}
\toprule
\rowcolor{headgray}
\textbf{Benchmark} & \textbf{\#P} & \textbf{W} & \textbf{U} & \textbf{Dep.} & \textbf{Depth} & \textbf{Obj.} & \textbf{Subj.} & \textbf{ST-DAG} & \textbf{Diag.} \\
\midrule
VBench \cite{vbench} & 946 & 7.7 & 1.7 & 0.1 & \na & \cmark & \xmark & \xmark & \xmark \\
VBench++ \cite{huang2025vbench++} & 3330 & 8.8 & 1.7 & 0.2 & \na & \cmark & \xmark & \xmark & \xmark \\
VBench-2.0 \cite{vbench2.0} & 1230 & 20.2 & 3.8 & 1.4 & \na & \cmark & \cmark & \xmark & \xmark \\
T2V-CompBench \cite{sun2025t2vcompbench} & 1400 & 10.4 & 1.6 & 0.4 & \na & \cmark & \xmark & \xmark & \xmark \\
TC-Bench \cite{feng2024tcbench} & 270 & 12.0 & 1.4 & 1.2 & \na & \cmark & \xmark & \xmark & \xmark \\
VMBench \cite{ling2025vmbench} & 1050 & 26.3 & 4.7 & 1.9 & \na & \na & \na & \xmark & \xmark \\
T2VWorldBench \cite{chen2026t2vworldbench} & 1260 & 11.2 & 1.6 & 0.4 & \na & \cmark & \xmark & \xmark & \xmark \\
ChronoMagic \cite{yuan2024chronomagicbench} & 1649 & 45.2 & 8.4 & 4.8 & \na & \na & \na & \xmark & \xmark \\
GenAI-Bench \cite{li2024genaibench} & 512 & 12.5 & 2.2 & 0.5 & \na & \xmark & \cmark & \xmark & \xmark \\
MJ-Video \cite{tong2025mjvideo} & 1085 & 46.5 & 7.8 & 2.9 & \na & \xmark & \cmark & \xmark & \xmark \\
\midrule
\rowcolor{oursblue}
\textbf{VGIF-Bench (Ours)} & \textbf{223} & \textbf{78.2} & \textbf{11.3} & \textbf{6.5} & \textbf{4.9} & \textbf{\cmark} & \textbf{\cmark} & \textbf{\cmark} & \textbf{\cmark} \\
\bottomrule
\end{tabular*}

\vspace{2pt}
\parbox{\textwidth}{\raggedright\scriptsize
\textbf{Abbr.} W: average words; U: atomic units; Dep.: estimated dependencies; Obj.: objective assessment; Subj.: subjective assessment; ST-DAG: explicit spatio-temporal directed acyclic graph; Diag.: diagnostic evaluation. U and Dep. are uniformly estimated from prompt text for fair comparison. Depth is reported only when an explicit graph structure is available. For VGIF-Bench, the explicit ST-DAG contains 16.4 nodes and 17.7 edges per prompt on average.}
\end{table*}

%
%
%

\section{Related Works}

\subsubsection{Video Generation Models} VGMs have evolved from early generative models to large-scale diffusion and diffusion-transformer systems~\cite{VAE,goodfellow2014generative,yan2021videogpt}. Recent models have greatly improved visual fidelity, motion quality, and temporal consistency~\cite{wan2025wan,team2025kling}, enabling increasingly realistic and cinematic video synthesis from open-ended text prompts. These advances are largely driven by stronger spatio-temporal modeling, larger training corpora, and more expressive generative backbones. However, improved visual realism does not necessarily imply faithful instruction following. A video may look plausible at the frame or clip level while still omitting later constraints, confusing object states, or breaking the causal relation between events.

\subsubsection{Video Generation Benchmarks and Metrics}
VGM evaluation has progressed from automatic metrics such as FVD~\cite{unterthiner2018towardsFVD} and CLIP-based similarity~\cite{clipscore,tang2024clip} to comprehensive benchmark suites. Recent works evaluate video quality and consistency across multiple dimensions~\cite{vbench,vbench2.0,huang2025vbench++}, incorporate human-aligned or preference-oriented assessment~\cite{evalcrafter}, study compositionality and text-video alignment~\cite{sun2025t2vcompbench}, or explore physical reasoning, world knowledge, long-context prompts, and MLLM-based judging~\cite{chen2026t2vworldbench,wang2026cinetechbench,wang2026detailverifybenchbenchmarkdensehallucination,yang2026evalversepipelineawareexpertcalibratedbenchmarking,tang2026revisiting,tang2026endogenous}. These efforts have substantially broadened the scope of video generation evaluation, covering visual quality, temporal consistency, motion realism, text-video alignment, and user preference. Nevertheless, most protocols still treat prompts primarily as flat text, and their evaluation targets are often defined at the dimension or video level. As a result, they provide limited information about which atomic constraints are satisfied, which prerequisite events are missing, and how an early failure affects downstream state changes or causal outcomes. 

\section{VGIF-Score}

\subsection{Framework Overview}

Given a text prompt $p$ and a generated video $x$, VGIF-Score evaluates how faithfully $x$ follows the instruction in $p$. As shown in Figure~\ref{fig:vgif_pipeline}, it consists of two complementary branches: \emph{objective completion} and \emph{subjective satisfaction}.

The objective branch uses an LLM to parse $p$ into a \textbf{Spatio-Temporal Directed Acyclic Graph (ST-DAG)}. Based on this graph, we construct dependency-aware QA pairs and use a VLM evaluator to answer them against the generated video. The QA accuracy yields an objective completion score.

The subjective branch uses an LLM to generate an \textbf{instruction-conditioned AutoRubric} tailored to the specific prompt $p$, each rubric dimension produces scoring criteria and anchor descriptions.  The VLM evaluator rates cinematography, visual purity, motion smoothness, and physics adherence on a 1--5 scale, normalized to $[0,1]$, and equally weighted. The final VGIF-Score combines the two branches with equal weights.

\subsection{ST-DAG-Based Objective Completion}

We represent each prompt $p$ as a \textbf{Spatio-Temporal Directed Acyclic Graph}:
\begin{equation}
\mathcal{G}(p)=(\mathcal{V},\mathcal{E}), \quad
\mathcal{Q}(p)=\{(q_i,a_i)\}_{i=1}^{N},
\end{equation}
where each node $v\in\mathcal{V}$ denotes an atomic semantic unit and each directed edge $e\in\mathcal{E}$ denotes a dependency relation. The node set covers six types---\emph{entity}, \emph{attribute}, \emph{location}, \emph{action}, \emph{state}, and \emph{causal}---progressing from static scene elements to dynamic events and their consequences. Edge types include \emph{solid dependencies} (compositional prerequisites) and \emph{causal dependencies} (consequence relations). Based on $\mathcal{G}(p)$, we derive $N$ dependency-aware QA pairs, where $q_i$ is a binary question associated with a graph node and $a_i$ is its expected answer.

\textbf{Dependency-aware evaluation.}
Given video $x$, a VLM evaluator answers every question independently. Before
crediting a node, we verify that its dependency expression is satisfied.
Dependencies follow the logical connectives specified in the ST-DAG
annotation: conjunctive (AND) dependencies require all upstream nodes to be
correct, while disjunctive (OR) dependencies require at least one. If the
dependency expression evaluates to false, the node is marked incorrect
regardless of its own answer, and this failure propagates to all downstream
nodes along the dependency chain.

Formally, let $\hat{a}_i$ be the VLM's answer for question $q_i$, and let
$\mathrm{dep}(i)$ denote its dependency expression over upstream QA indices.
The per-node correctness is defined recursively:
\begin{equation}
c_i =
\mathbf{1}[\hat{a}_i = a_i]
\;\wedge\;
\mathrm{eval}\bigl(\mathrm{dep}(i),\; \{c_j\}_{j<i}\bigr),
\end{equation}
where $\mathrm{eval}(\cdot)$ evaluates the Boolean dependency expression
(supporting AND, OR, and parentheses) against previously computed correctness
values. The objective completion score is:
\begin{equation}
S_{\mathrm{obj}}(p,x)=\frac{1}{N}\sum_{i=1}^{N} c_i.
\end{equation}

\subsection{Instruction-Conditioned AutoRubric}

While the objective branch verifies whether individual semantic units are realized, it does not capture holistic perceptual qualities that affect user satisfaction. We therefore introduce an \textbf{instruction-conditioned AutoRubric} that evaluates four complementary dimensions:

\begin{itemize}
\item \textbf{Cinematography}: whether camera work, composition, lighting, and pacing match the narrative tone specified in the prompt.
\item \textbf{Visual purity}: whether the video contains \emph{only} elements specified in the prompt, with no extraneous objects, identity drift, or artifacts.
\item \textbf{Motion smoothness}: whether movements and interactions are temporally smooth, continuous, and free of jitter or freezing.
\item \textbf{Physics adherence}: whether prompt-specified interactions and state changes follow plausible physical behavior.
\end{itemize}

Crucially, the rubric is not generic: for each prompt, the LLM generates \emph{prompt-specific scoring criteria} and \emph{score anchor descriptions} for every dimension, so that the evaluator judges the video against what was actually requested rather than an abstract quality standard.

For each dimension $k\in\{1,2,3,4\}$, the evaluator assigns a raw score $r_k(p,x)\in\{1,2,3,4,5\}$. We normalize and aggregate:
\begin{equation}
\tilde{r}_k(p,x)=\frac{r_k(p,x)-1}{4}, \quad
S_{\mathrm{rubric}}(p,x)=\frac{1}{4}\sum_{k=1}^{4}\tilde{r}_k(p,x).
\end{equation}

\subsection{Final Score}

The final VGIF-Score combines objective completion and subjective satisfaction:
\begin{equation}
S_{\mathrm{VGIF}}(p,x)=
\frac{1}{2}\,S_{\mathrm{obj}}(p,x)+
\frac{1}{2}\,S_{\mathrm{rubric}}(p,x).
\end{equation}
For a benchmark of $M$ prompt-video pairs, the overall score is:
\begin{equation}
\mathrm{VGIF\text{-}Score}
=
\frac{1}{M}\sum_{m=1}^{M}S_{\mathrm{VGIF}}(p_m,x_m).
\end{equation}

\section{VGIF-Bench}

\begin{figure*}[t]
    \centering
    \includegraphics[width=\textwidth]{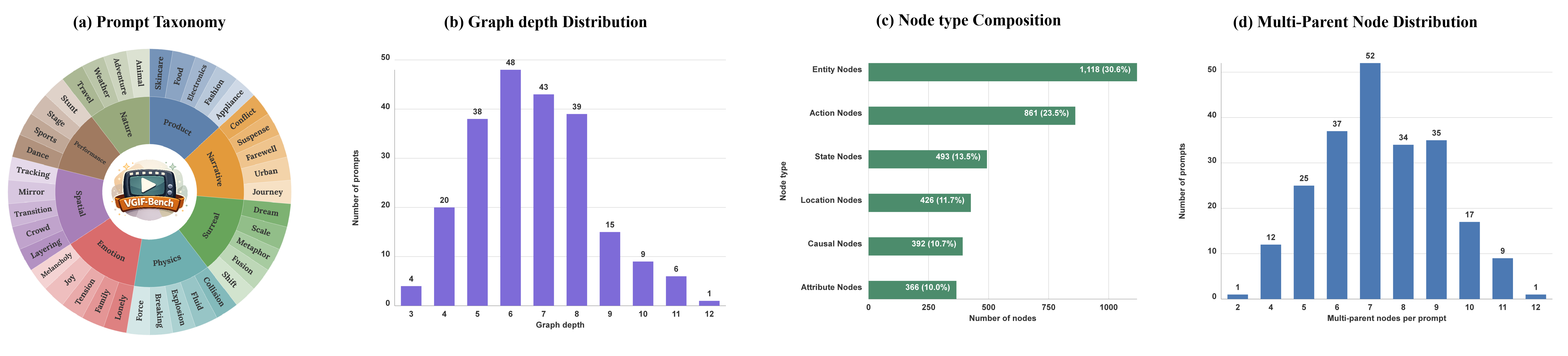}
    \vspace{-4pt}
\caption{
Overview of VGIF-Bench. The figure presents (a) the hierarchical prompt taxonomy, (b) graph depth distribution, (c) ST-DAG node-type composition, and (d) multi-parent node distribution. These statistics illustrate the coverage and structural complexity of the benchmark.}
    \vspace{-6pt}
    \label{fig:vgif_bench}
\end{figure*}

VGIF-Bench is designed to evaluate whether video generation models can faithfully execute long, structurally entangled instructions rather than merely depict isolated objects or short event fragments. Unlike prior benchmarks that treat prompts as flat text, VGIF-Bench represents each prompt as an explicit spatio-temporal directed acyclic graph (ST-DAG), aligns graph nodes with dependency-aware QA, and complements structural verification with instruction-conditioned autorubric scoring. This design makes the benchmark not only more challenging, but also substantially more interpretable and diagnostic.

\subsection{Benchmark Construction}

We construct \textbf{VGIF-Bench} through a largely automated pipeline with human verification, prioritizing structural instruction complexity over benchmark scale. Starting from a hierarchical taxonomy of video-generation scenarios, GPT-5.2 \cite{gpt52} drafts each benchmark sample together with (1) a long-form prompt, (2) its ST-DAG decomposition, (3) dependency-aware QA pairs, and (4) instruction-conditioned autorubric specifications. Automatic validation then enforces schema consistency, dependency validity, duration normalization, and sample deduplication, after which human annotators verify graph semantics, QA answerability, and rubric alignment. This design allows the expensive human effort to focus on verification and correction, rather than writing all prompts and evaluation items from scratch.

The final benchmark contains \textbf{223 prompts}, \textbf{3,445 dependency-aware QA pairs}, and \textbf{892 autorubric dimension specifications}, yielding approximately \textbf{4.3K fine-grained evaluation items} across the objective and subjective branches. Across the benchmark, the ST-DAG annotations contain \textbf{3,656 nodes} and \textbf{3,940 edges} in total, corresponding to \textbf{16.4 nodes} and \textbf{17.7 edges} per prompt on average.

\subsection{Benchmark Structure and Distribution}

Figure~\ref{fig:vgif_bench} summarizes the coverage and structural complexity of VGIF-Bench. As shown in Figure~\ref{fig:vgif_bench}(a), the benchmark spans \textbf{8 macro categories} and \textbf{38 subcategories}, covering product, narrative, spatial, emotional, physical, performative, natural, and surreal scenarios. This taxonomy is intended to capture both common real-world generation requests and compositionally difficult prompts that require coordinated multi-step execution.

Figure~\ref{fig:vgif_bench}(b) shows the distribution of graph depth across prompts. Most samples exhibit non-trivial hierarchical structure, with the mass concentrated in the mid-to-high depth range rather than near-flat dependency chains. This indicates that VGIF-Bench is challenging not only because prompts are long, but because many instructions require multi-stage semantic execution under explicit dependency constraints.

Figure~\ref{fig:vgif_bench}(c) presents the composition of ST-DAG node types. While entity and action nodes form the dominant backbone, VGIF-Bench also contains substantial numbers of state, location, attribute, and causal nodes. This makes temporal evolution and inter-event dependency first-class evaluation targets rather than incidental byproducts of prompt wording.

Figure~\ref{fig:vgif_bench}(d) highlights the prevalence of non-linear dependency patterns. Many prompts contain multiple nodes whose realization depends on several upstream conditions simultaneously, rather than simple left-to-right chains. Such multi-parent structures are important because they enable fine-grained diagnosis of failure propagation and reveal whether a model can maintain coherent execution across intertwined semantic constraints.

Taken together, these properties make VGIF-Bench difficult not because it is large, but because successful generation requires coordinated realization of entities, actions, states, and causal outcomes under explicit structural dependencies. These properties directly motivate our evaluation design: objective QA localizes constraint-level failures, while AutoRubric captures perceptual degradation caused by broken event execution.

\section{Evaluation}

\subsection{Experimental Setup}

\textbf{Models.}
We evaluate 14 representative VGMs, including five proprietary models with undisclosed parameters (Kling-V3, Seedance-2.0, Wan-2.7, ViduQ3-Turbo, and PixVerse-V6) and nine open-source models with publicly released weights: LTX-2.0 (19B) \cite{hacohen2026ltx2.0}, Wan2.2-A14B (27B total, 14B active) \cite{wan2025wan}, HyVideo-1.5 (HunyuanVideo-1.5, 8.3B) \cite{wu2025hunyuanvideo1.5}, LongCat-Video (13.6B) \cite{team2025longcat}, Mochi-1 (10B) \cite{genmo2024mochi}, CogVideoX-1.5 (5B) \cite{yang2024cogvideox}, MAGI-1 (4.5B) \cite{ai2025magi1autoregressivevideogeneration}, URSA (1.7B) \cite{deng2025ursa}, and InfinityStar (8B) \cite{InfinityStar}. 
This model pool covers both diffusion-transformer variants and autoregressive video generation, and spans a broad range of parameter scales from compact 1.7B models to large-scale 27B systems, providing a comprehensive testbed for evaluating spatio-temporal instruction following.

\textbf{Evaluation model and protocol.}
We use \textbf{Gemini-3.1-Pro} \cite{gemini31} as the unified VLM evaluator for both QA-based objective completion and AutoRubric-based subjective assessment.
For each prompt-video pair, VGIF-Score is computed at the video level and then averaged over the benchmark.
For dimension-wise analysis, such as entity, action, and causal relation, we aggregate QA accuracy across all questions belonging to the corresponding semantic dimension.
\begin{table*}[t]
\caption{Main results on VGIF-Bench. The first-ranked result is highlighted in blue, and the second-ranked result in green. Columns are organized as Entity, Attribute, Location, Action, State, Causal, Objective Score, Cinematography, Visual Purity, Motion Smoothness, Physics Adherence, Subjective Score, and VGIF-Score.}
\label{tab:main_results}
\centering
\scriptsize
\renewcommand{\arraystretch}{1.10}
\setlength{\heavyrulewidth}{0.9pt}
\setlength{\lightrulewidth}{0.45pt}
\setlength{\cmidrulewidth}{0.45pt}
\setlength{\tabcolsep}{2.2pt}

\begin{adjustbox}{max width=\textwidth}
\begin{tabular}{lccccccccccccc}
\toprule
\rowcolor{headgray}
\multirow{2}{*}{\textbf{Model}}
& \multicolumn{7}{c}{\textbf{Objective}}
& \multicolumn{5}{c}{\textbf{Subjective}}
& \multirow{2}{*}{\textbf{VGIF}} \\
\cmidrule(lr){2-8}\cmidrule(lr){9-13}
\rowcolor{headgray}
& Ent. & Attr. & Loc. & Act. & Sta. & Cau. & Obj.
& Cin. & Pur. & Mot. & Phy. & Sub. & \\
\midrule

\rowcolor{groupgray}
\multicolumn{14}{l}{\textit{\textbf{\textcolor{commblue}{Commercial Models}}}} \\
Kling-V3      & 71.07 & 57.36 & 75.14 & 20.73 & 12.60 & \best{4.21} & 42.18 & \second{50.58} & \best{74.35} & \second{49.06} & \second{37.58} & \second{52.89} & 46.30 \\
Seedance-2.0  & 70.18 & 54.01 & 75.00 & 19.34 & 11.65 & 2.98 & 40.96 & \best{55.67} & \second{71.52} & \best{55.76} & \best{43.41} & \best{56.59} & \best{47.59} \\
Wan-2.7       & \best{76.46} & \best{70.91} & \best{82.73} & 22.12 & \second{13.21} & 3.46 & \second{46.29} & 45.88 & 64.89 & 42.35 & 33.21 & 46.58 & 46.44 \\
ViduQ3-Turbo  & 74.35 & 66.07 & 78.73 & \best{35.57} & 10.95 & \second{3.68} & 44.76 & 44.93 & 67.62 & 46.19 & 34.35 & 48.27 & 45.35 \\
PixVerse-V6   & \second{75.41} & \second{66.97} & \second{81.22} & \second{25.91} & \best{15.50} & \best{4.21} & \best{46.73} & 48.79 & 66.73 & 44.84 & 35.34 & 48.93 & \second{47.18} \\

\midrule
\rowcolor{groupgray}
\multicolumn{14}{l}{\textit{\textbf{\textcolor{osgreen}{Open-Source Models}}}} \\
LTX-2.0      & 57.38 & 45.65 & 61.88 & 8.95 & 4.34 & 0.53 & 31.06 & \second{40.00} & \second{57.49} & \second{44.48} & \second{36.59} & \second{44.64} & 36.50 \\
Wan2.2-A14B  & \best{69.33} & \best{60.06} & \best{74.31} & \best{14.37} & \best{8.88} & \best{1.84} & \best{39.48} & 39.10 & 56.77 & 37.31 & 28.61 & 40.45 & \best{39.96} \\
HyVideo-1.5  & 59.79 & 50.15 & 65.75 & \second{12.37} & \second{6.20} & \second{0.79} & 33.76 & \best{44.30} & \best{60.36} & \best{47.53} & \best{37.40} & \best{47.40} & \second{39.18} \\
LongCat-Video & \second{64.42} & 55.26 & 66.02 & 11.07 & 5.79 & 0.53 & 35.27 & 39.01 & 52.74 & 42.06 & 32.02 & 41.46 & 38.47 \\
Mochi-1      & 56.03 & 50.45 & 65.19 & 9.66 & 5.37 & 0.26 & 31.76 & 35.53 & 52.65 & 33.15 & 28.07 & 37.35 & 33.28 \\
CogVideoX-1.5 & 52.75 & 51.65 & 63.26 & 9.54 & 4.13 & 0.00 & 30.37 & 28.43 & 45.74 & 31.21 & 25.47 & 32.71 & 31.54 \\
MAGI-1       & 44.74 & 36.34 & 59.94 & 3.30 & 2.27 & 0.00 & 24.63 & 24.04 & 41.97 & 26.28 & 23.14 & 28.86 & 26.74 \\
URSA         & 51.69 & 49.25 & 60.22 & 5.18 & 2.89 & 0.00 & 28.33 & 29.60 & 41.08 & 34.80 & 27.09 & 33.14 & 30.92 \\
InfinityStar & 62.68 & \second{59.46} & \second{72.10} & 11.07 & 2.89 & 0.00 & \second{35.33} & 38.61 & 52.24 & 44.38 & 31.24 & 35.52 & 35.43 \\

\bottomrule
\end{tabular}
\end{adjustbox}
\end{table*}

\begin{table}[t]
\caption{VGIF-Score by scenario category. The first-ranked result is highlighted in blue, and the second-ranked result in green.}
\label{tab:category_results}
\centering
\scriptsize
\renewcommand{\arraystretch}{1.10}
\setlength{\heavyrulewidth}{0.9pt}
\setlength{\lightrulewidth}{0.45pt}
\setlength{\tabcolsep}{3.5pt}

\begin{tabular}{lcccccccc}
\toprule
\rowcolor{headgray}
\textbf{Model} & Product & Narrative & Surreal & Physics & Emotion & Spatial & Performance & Nature \\
\midrule

\rowcolor{groupgray}
\multicolumn{9}{l}{\textit{\textbf{\textcolor{commblue}{Commercial Models}}}} \\
Kling-V3      & 46.13 & \second{43.12} & 44.85 & 43.25 & 50.39 & 48.78 & 49.31 & 45.46 \\
Seedance-2.0  & \second{47.26} & \best{46.14} & \second{48.82} & 40.93 & \second{53.18} & \second{49.66} & 49.21 & 46.83 \\
Wan-2.7       & \best{49.77} & 43.03 & \best{53.17} & \best{50.62} & \best{65.48} & \best{58.20} & \best{61.60} & \best{58.06} \\
ViduQ3-Turbo  & 46.45 & 38.63 & 42.77 & 42.61 & 51.47 & 46.50 & \second{50.06} & 45.97 \\
PixVerse-V6   & {46.20} & 42.74 & 46.05 & \second{44.17} & 50.24 & 47.24 & 48.29 & \second{54.59} \\

\midrule
\rowcolor{groupgray}
\multicolumn{9}{l}{\textit{\textbf{\textcolor{osgreen}{Open-Source Models}}}} \\
LTX-2.0      & 33.00 & \second{34.78} & 34.65 & 34.03 & 37.70 & \second{40.18} & 38.79 & 40.20 \\
Wan2.2-A14B  & \second{38.16} & \best{37.65} & \best{41.59} & \best{39.57} & 32.23 & 39.96 & \second{39.95} & \best{47.58} \\
HyVideo-1.5  & \best{40.43} & 29.47 & \second{35.06} & 35.62 & \best{47.67} & 39.36 & \best{43.47} & \second{45.26} \\
LongCat-Video & 37.73 & 34.09 & 33.23 & \second{36.56} & \second{44.95} & 40.12 & 38.95 & 43.86 \\
Mochi-1      & 33.87 & 27.40 & 34.39 & 30.42 & 39.44 & 34.72 & 32.14 & 34.80 \\
CogVideoX-1.5 & 28.81 & 29.87 & 29.71 & 26.00 & 36.37 & 35.92 & 30.43 & 30.43 \\
MAGI-1       & 27.55 & 24.46 & 25.87 & 28.55 & 29.39 & 29.39 & 24.06 & 26.80 \\
URSA         & 33.19 & 25.23 & 32.70 & 30.64 & 31.97 & 29.61 & 30.66 & 34.12 \\
InfinityStar & 33.44 & 27.06 & 32.93 & 33.26 & 41.65 & \best{40.47} & 34.46 & 42.02 \\

\bottomrule
\end{tabular}
\end{table}

\subsection{Main Results}

Table~\ref{tab:main_results} reports the main results on VGIF-Bench.

\textbf{Overall performance.} Proprietary models achieve an average VGIF-Score of 46.57, compared with 34.67 for open-source models, showing a clear but not overwhelming gap. Seedance-2.0 obtains the highest overall VGIF-Score among proprietary models, while Wan2.2-A14B leads the open-source group. Nevertheless, even the best-performing models remain far from fully satisfying VGIF-Bench, indicating that spatio-temporal instruction following is still a challenging capability beyond visual fidelity. 

\textbf{Objective vs. subjective gap.} High subjective quality does not necessarily imply strong objective completion. Several models obtain reasonable visual purity or motion scores but remain weak on action, state, and causal dimensions. This discrepancy supports the need for a dual-branch metric: subjective quality alone may overestimate visually plausible but semantically incomplete videos, while objective QA alone cannot capture perceptual degradation.

\textbf{Causal reasoning bottleneck.} Causal instruction following is the most difficult objective dimension across nearly all models. Even the strongest commercial models obtain causal scores below 5, while multiple open-source models are close to zero. This indicates that current VGMs still struggle to bind events into coherent cause-effect chains, rather than simply rendering local objects or short actions. Importantly, causal failures are not isolated errors: once a triggering action or prerequisite state is missed, downstream state changes and causal outcomes often become impossible to realize. This explains why causal scores are substantially lower than entity or location scores, and motivates the dependency-aware short-circuit design of VGIF-Score.

\subsection{Category-wise Analysis}

We further analyze performance across scenario categories in Table~\ref{tab:category_results}. 

\textbf{Scenario difficulty.} Different categories expose distinct model weaknesses. Emotion, performance, and nature scenarios tend to receive higher scores, partly because they often rely more on appearance, style, or short-range motion. In contrast, narrative and physics-related scenarios are more challenging because they require richer temporal evolution, state transitions, and causal dependencies. This trend is consistent with the structural statistics in Figure~\ref{fig:vgif_bench}, where deeper dependency chains and multi-parent structures are common sources of difficulty. 

\textbf{Model bias.} Models also exhibit category-specific biases. Some systems perform competitively on appearance-driven categories but degrade in structured or multi-entity interactions. For example, a model may generate visually appealing product or nature videos while failing to maintain coherent event progression in narrative or physics scenarios. Such results show that high visual quality alone does not guarantee robust generalization to dependency-rich instructions. 

\textbf{Implication.} The category-wise results suggest that future evaluation should not only report a single overall score, but also expose which types of instructions stress a model. A model optimized for visual appeal may rank highly on product or nature prompts, yet still fail on categories that require event-level reasoning. VGIF-Bench therefore provides a more diagnostic view of model capability by linking scenario-level performance with explicit structural properties.

\subsection{Structural Analysis}
\label{sec:structural}

VGIF-Bench's ST-DAG representation makes it possible to measure
\emph{where} in the prompt and at \emph{which} compositional depth
instruction following degrades.
We analyze accuracy along two orthogonal axes:
relative position of constraints in the prompt text, and
dependency depth in the ST-DAG.

\begin{figure}[!ht]
\centering
\includegraphics[width=\textwidth]{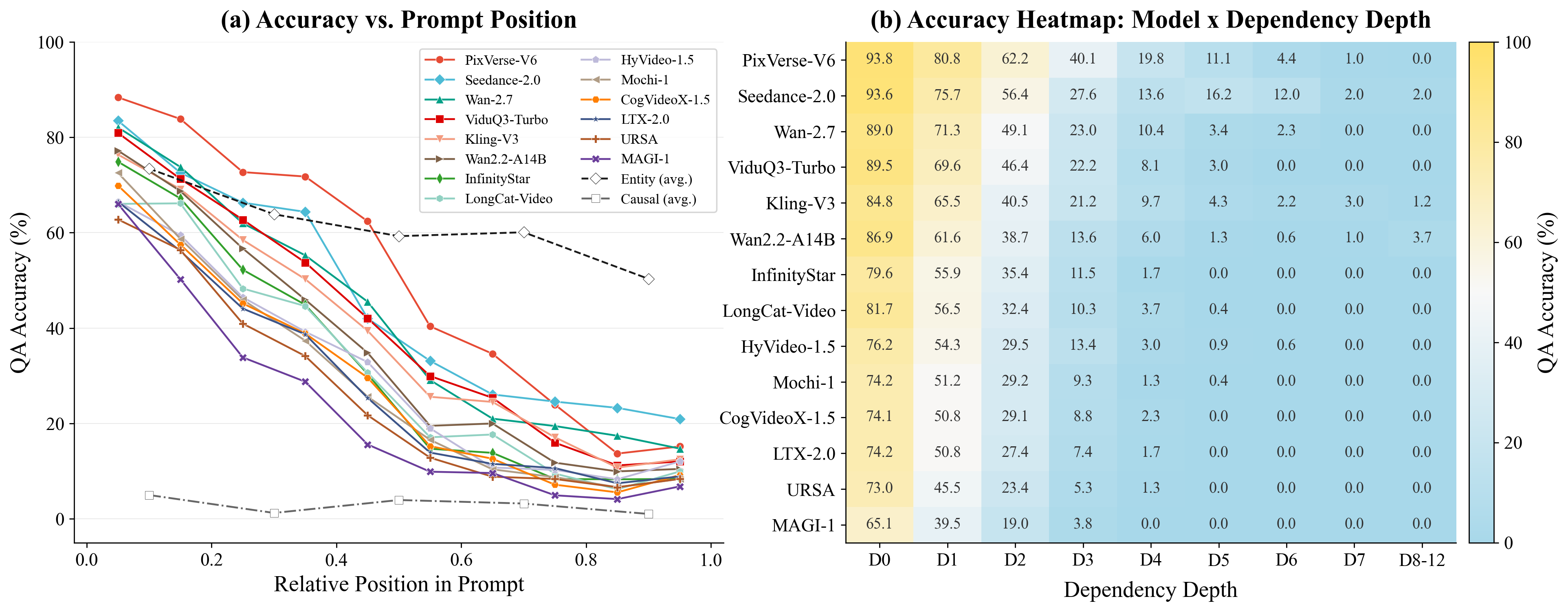}
\caption{\textbf{Structural factors governing instruction-following accuracy.}
(a) QA accuracy vs.\ relative position in the prompt.
Accuracy drops from 67.9\% (first 20\%) to 10.1\% (final 20\%), a $6.7\times$ decline
universal across all 14 VGMs.
(b) Heatmap of accuracy by model and dependency depth.
Accuracy decreases monotonically with depth; depths 8--12 are merged.
All values to one decimal place.}
\label{fig:structural}
\end{figure}

\textbf{Prompt position.}
Fig.~\ref{fig:structural}a reveals a strong recency bias: averaged across all 14 models,
the accuracy falls from 67.9\% for constraints appearing in the first 20\% of the prompt
to 48.9\% at 0.2--0.4, 27.3\% at 0.4--0.6, 14.9\% at 0.6--0.8, and 10.1\% in the
final 20\%---a $6.7\times$ decline.
Even PixVerse-V6 drops from 85.5\% at early positions to 14.4\% at late positions.
Entity questions retain 50.3\% accuracy at late positions,
while causal questions drop to 1.0\% past the midpoint,
showing that position sensitivity is most severe for semantically complex constraints.

\textbf{Dependency depth.}
Fig.~\ref{fig:structural}b shows that precision decreases monotonically with ST-DAG depth: averaged across all 14 models,
80.6\% at depth~0 (independent questions), 58.5\% at depth~1, 36.4\% at depth~2,
15.1\% at depth~3 and 5.6\% at depth~4---an average drop of $\sim$19\% per level.
Beyond depth~4, accuracy falls below 3\% for all 14 models.
PixVerse-V6 retains 62.2\% at depth~2 while MAGI-1 drops to 19.0\%,
yet all models converge to near-zero beyond depth~4.
The depth~0$\rightarrow$1 decline (80.6\%$\rightarrow$58.5\%, a 22.1\% drop)
quantifies the immediate cost of even a single dependency.

\textbf{Interaction.}
Position and depth effects compound multiplicatively:
a causal constraint appearing late in the prompt and sitting at depth~$\ge$3
faces near-zero success probability across all models.
Together, these two orthogonal axes---temporal attention decay and compositional
reasoning depth---explain the dominant share of the instruction-following gap.

\subsection{Diagnostic Analysis}

\label{sec:casestudy}

\begin{figure}[t]
\centering
\includegraphics[width=\linewidth]{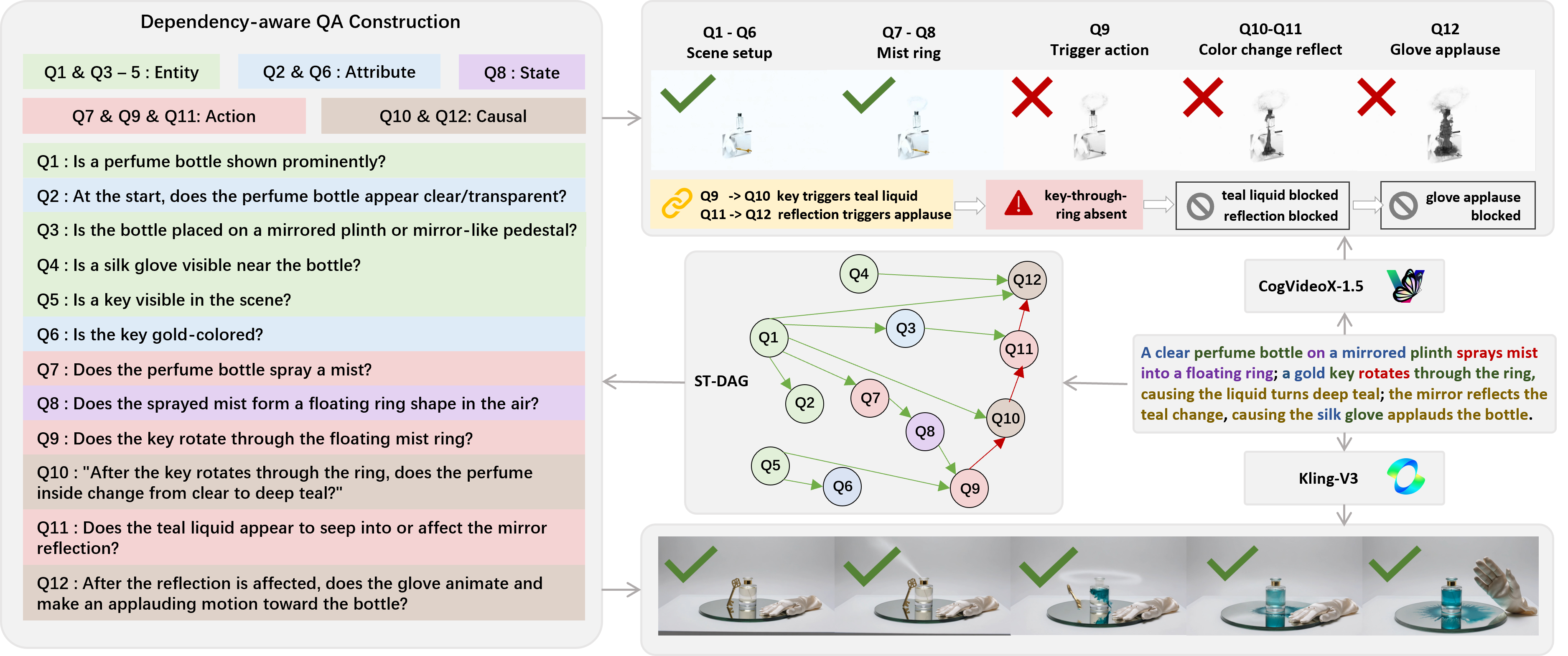}
\caption{\textbf{Dependency-aware causal chain diagnosis.}
Kling-V3 executes the full perfume transformation chain (12/12 QA, 2/2 causal).
CogVideoX-1.5 preserves the local scene and early mist formation (q1--q8), but misses the 
trigger action q9; the downstream causal nodes q10 and q12 fail under dependency-aware 
evaluation.}
\label{fig:casestudy}
\end{figure}

To illustrate how structural factors interact with both objective and subjective evaluation, Figure~\ref{fig:casestudy} presents a side-by-side diagnosis of Kling-V3 and CogVideoX-1.5 on the same VGIF-Bench prompt. The prompt describes a surreal perfume advertisement with an explicit causal chain: the bottle sprays mist, the mist forms a floating ring, a gold key rotates through the ring, the liquid shifts color, the mirror reflection changes, and a silk glove animates to applaud.

\textbf{Objective diagnosis.}
Kling-V3 executes the full dependency chain, answering all 12 dependency-aware QA pairs correctly, including both causal questions. In contrast, CogVideoX-1.5 satisfies the local scene and early state constraints but fails at the bridge action where the key should rotate through the mist ring. Because this action gates downstream causal and state nodes, the short-circuit mechanism propagates the single failure to later constraints. The ST-DAG therefore localizes the failure to the causal transition edge, rather than only reporting a flat aggregate accuracy.

\subsection{Human Validation and Alignment}

We randomly sampled 200 generated videos from VGIF-Bench and collected annotations from three human annotators per sample. Human annotators answer the same ST-DAG QA pairs, score videos with the same AutoRubric criteria, and provide overall ratings. We report Cohen's $\kappa$~\cite{cohen1960coefficient} for categorical QA agreement and Spearman rank correlation~\cite{spearman1961proof} for rating-based scores.
Table~\ref{tab:human_validation} summarizes two aspects of validation.
First, Gemini-3.1-Pro shows strong consistency with human annotations under the same evaluation protocol, supporting its effectiveness as the VLM evaluator.
Second, the objective branch better aligns with human completion judgments, while AutoRubric better aligns with subjective satisfaction.
The final VGIF-Score achieves the highest correlation with human overall ratings, demonstrating that combining structural correctness and perceptual quality provides a more comprehensive evaluation signal.

\begin{table}[t]
\centering
\caption{Human validation and alignment.}
\label{tab:human_validation}
\scriptsize
\renewcommand{\arraystretch}{1.12}
\setlength{\tabcolsep}{4.4pt}
\begin{tabular}{lccc}

\multicolumn{4}{c}{\textbf{(a) Evaluator Effectiveness}} \\
\midrule
\rowcolor{headgray}
\textbf{Signal} & \textbf{Human Reference} & \textbf{Statistic} & \textbf{Value} \\
\midrule
ST-DAG QA      & QA labels          & Agreement      & $96.3\%$ \\
ST-DAG QA      & QA labels          & Cohen's $\kappa$ & 0.92 \\
AutoRubric     & Rubric scores      & Spearman $\rho$ & 0.83 \\
VGIF-Score     & Human-derived VGIF & Spearman $\rho$ & 0.87 \\
\midrule[0.6pt]

\multicolumn{4}{c}{\textbf{(b) Human Alignment}} \\
\midrule
\rowcolor{headgray}
\textbf{Automatic Score} & \textbf{Human-Comp.} & \textbf{Human-Sat.} & \textbf{Human-Overall} \\
\midrule
Objective Score  & 0.78 &0.52 & 0.65 \\
AutoRubric Score & 0.41 & 0.81 & 0.72 \\
VGIF-Score       &0.71 & 0.83 & \textbf{0.89} \\
\bottomrule
\end{tabular}
\end{table}

\section{Conclusion}

We introduced VGIF-Score, an interpretable and diagnostic framework for evaluating spatio-temporal instruction following in video generation. By combining ST-DAG-based objective completion with instruction-conditioned AutoRubric assessment, VGIF-Score measures both structural correctness and perceptual satisfaction while localizing where failures occur. We further built VGIF-Bench, a dependency-rich benchmark of 223 long-form prompts and approximately 4.3K fine-grained evaluation items, designed to evaluate multi-entity interactions, state transitions, and causal event chains. Experiments on 14 proprietary and open-source VGMs show that current models still struggle with deep instruction following, especially under causal chains, deep dependency structures, and late-position constraints. Human validation further supports the reliability of the VLM evaluator and the necessity of the dual-branch design. We hope our work can support more diagnostic evaluation and guide future video generation models toward stronger semantic and causal instruction following.

\section{Acknowledgements}
This work was supported by the National Nature Science Foundation of China (Grant U23B2052, 62225601, 62476029) and the Beijing Key Laboratory of Multimodal Data Intelligent Perception and Governance.

\bibliographystyle{splncs04} 
\bibliography{refs}          

\end{document}